\documentclass[12pt]{article}
\usepackage[margin=1in]{geometry}
\usepackage[T1]{fontenc}
\usepackage[utf8]{inputenc}
\usepackage[american]{babel}
\usepackage{amsmath,amssymb}
\usepackage{graphicx}
\usepackage{booktabs}
\usepackage{multirow}
\usepackage{caption}
\usepackage{subcaption}
\usepackage{float}
\usepackage{achemso}
\setkeys{acs}{articletitle=true}
\usepackage{authblk}

\usepackage[colorlinks=true, allcolors=blue]{hyperref}
\graphicspath{{figures/}}
\title{Efficient Patch Based Anomaly Detection Fused with Diffusion Driven Generative Modeling for Semiconductor Wafer Bin Map Open Set Anomaly Detection}
\author[1,*]{Limon Bin Hossain}
\author[1]{Md Sadib Rahman Ananta}

\affil[1]{Department of Industrial and Production Engineering, Bangladesh University of Engineering and Technology (BUET), Dhaka, 1000, Bangladesh}

\affil[*]{Corresponding author: \href{mailto:mdlimonbinhossain@gmail.com}{mdlimonbinhossain@gmail.com}}
\date{} 

\begin{document}
\maketitle
\noindent{\bf Abstract}

Spatial defect signatures on wafer bin maps (WBMs) trace yield loss to specific
process faults, yet supervised classifiers recognize only the defect types seen during training, and one-class detectors built on a single mechanism tend to
capture either local structural deviations or global distributional violations,
but rarely both. This work proposes a hybrid one-class framework that couples a
patch-based student--teacher detector (EfficientAD) with a denoising diffusion
probabilistic model (DDPM) used for partial-diffusion reconstruction, and fuses
their percentile-calibrated scores through a fixed convex combination. Trained
on only 700 normal wafers from the WM-38K mixed-type dataset and evaluated on
18,658 held-out wafers, the fused detector reached an AUROC of 0.9985 and reduced
misclassifications from 852 (DDPM) and 1,412 (EfficientAD) to 618, with all
pairwise differences significant at $p < 0.001$. Beyond aggregate accuracy, the
analysis shows that the gain arises from weakly overlapping errors between the
two modules, yet fixed-weight fusion recovers only 40--70\% of the correction
available to an oracle selector. Under the benchmark's inverted class balance,
average precision and F1 saturate, while the Matthews correlation coefficient and
negative predictive value expose unreliable normal predictions. Pixel-level maps
further show that strong image-level separability does not imply spatial
localization, and the diffusion module succeeds as a local density prior rather
than through global geometric reasoning. These findings motivate sample-adaptive
fusion and imbalance-aware evaluation of hybrid wafer anomaly detectors.

\vspace{0.2cm}

\section{Introduction}
\label{sec:introduction}

In semiconductor fabrication, each die on a wafer is tested, and results are recorded on a wafer bin map (WBM). The spatial patterns from these tests provide diagnostic insights, revealing issues like scratches and defects that can be traced to specific equipment or process steps \citep{nakazawa2018}. Unclassified signatures can lead to yield loss and production delays, underscoring the importance of not only detecting them but also quickly resolving them. Traditionally, defect recognition has been approached as a supervised multi-label classification problem, where convolutional networks excel under closed-set conditions \citep{wang2020}. However, these classifiers can only identify trained defect types and may fail on significantly different patterns. An alternative approach is to frame wafer defect detection as a one-class anomaly detection problem. This method primarily trains models on defect-free wafers \citep{batzner2024efficientad}, thereby flagging any significant deviation as anomalous. Two main families of one-class methods exist: one that compares a test wafer to a reference feature representation and another that models a learned data distribution. The structural approach uses a pretrained teacher network and a student network trained on defect-free images to identify anomalies by comparing patch-level features \citep{bergmann2020}. This has been further developed for patch-level feature distribution modeling \citep{defard2021}, one-class bottleneck reconstruction \citep{dengli2022}, and rapid inference in the EfficientAD architecture \citep{batzner2024}. While structural detectors effectively identify small anomalies, their patch-level focus may hinder the detection of patterns that emerge only across the entire image.

The second family of one-class methods addresses the limitation of patch-based modeling by analyzing the image as a whole. Generative models learn the distribution of normal images and use reconstruction quality to identify deviations. Denoising diffusion probabilistic models (DDPMs) are one such approach \citep{ho2020}, enhanced by learned variance schedules \citep{nichol2021} and deterministic implicit sampling \citep{song2021}. They adapt to reconstruction tasks by partially diffusing a test image and measuring the discrepancy between the input and its denoised version \citep{wyatt2022}. This approach captures the overall structure of an image, allowing it to detect distributional anomalies that patch-level methods may miss. However, global reconstruction methods are less sensitive to small, localized texture defects, and diffusion models using standard Gaussian noise can lose fine-grained local information during the diffusion process \citep{wyatt2022}.

The complementary limitations of structural and generative detectors motivate their combination in a hybrid framework by calibrating anomaly scores and applying fixed weights. However, ensemble methods do not always enhance performance through simple score averaging, especially if weaker detector errors correlate with stronger ones or contribute uninformative variation, which can dilute the reliable detector's contribution \citep{aggarwal2015}. An effective hybrid model requires a strategy that accounts for both the reliability and the correlation of its detectors. This study investigates the impact of detector combinations on anomaly detection performance, treating the fusion mechanism as a core aspect of the detection framework to systematically evaluate each detector's contributions and interactions.

These considerations raise three research questions that this study is designed to answer.

\begin{itemize}
  \item[\textbf{RQ1:}] Does fixed-weight score fusion of a patch-based structural anomaly detector and a diffusion-based generative anomaly detector improve wafer defect detection performance compared to the individual detectors, and when does fusion lead to performance degradation \citep{aggarwal2015}?
  
  To answer it, we train both detectors, fuse their calibrated scores with a fixed convex combination, and compare all three detectors under bootstrap-validated and McNemar-tested conditions.

  \item[\textbf{RQ2:}] How does the fused score relate geometrically to each detector's calibrated score across evaluation metrics, and is this relationship monotonic?
  
  To answer it, we examine the fusion mechanism using score distributions, receiver operating characteristic (ROC) curves, precision-recall curves, and calibration reliability diagrams for the structural, generative, and fused detectors.

  \item[\textbf{RQ3:}] What structural properties and diffusion-based scoring explain the resulting detector ranking, and what does this imply for the fusion and evaluation of hybrid anomaly detectors?
  
  To answer it, we inspect per-sample anomaly localization maps, the DDPM's generative sampling behavior, and its detection performance.
\end{itemize}

\section{Materials and Methods}
\label{sec:methods}

\subsection{Dataset and Preprocessing}
\label{sec:met_data}

WM-38K (MixedWM38) comprises 38,015 wafer bin maps with a resolution of $52\times52$\citep{wang2020}. Each map is an integer array over $\{0,1,2\}$ (blank, normal-die, broken-die), paired with an 8-bit label vector over eight defect primitives (center, donut, edge-loc, edge-ring, loc, near-full, scratch, random) that decodes to one of 38 classes. The mixing degree $m \in \{0,1,2,3,4\}$ (the sum of the label vector) has 1,000, 7,015, 13,000, 13,000, and 4,000 wafers at $m=0$ through $m=4$. Binary labels were set as $y = \mathbb{1}[m>0]$, giving 1,000 normal wafers (2.6\%) and 37,015 defective wafers (97.4\%), the inverse of the typical industrial base rate.

Each map was one-hot encoded into three binary channels (blank, normal-die, broken-die), zero-padded by six pixels on each side to $3\times64\times64$, and rescaled to $[-1,1]$ by $x \mapsto 2x-1$.

The 1,000 normal wafers were split 70\%/15\%/15\% into training (700), validation (150), and test (150); only the 700 training normals were used to fit any parameter. The 37,015 defective wafers, never used for training, were split 50\%/50\% between validation (18,507) and test (18,508). The validation partition (18,657 wafers) was used for percentile calibration and threshold selection (Section~\ref{sec:met_fusion}); the test partition (18,658 wafers) was reserved for the final metrics in Section~\ref{sec:results}.

\subsection{Structural Module: Patch-Based Student-Teacher Detection}
\label{sec:met_ead}

The structural branch follows EfficientAD, comparing a frozen pretrained teacher with a lightweight trainable student.\citep{batzner2024,bergmann2020} The teacher is an ImageNet-pretrained ResNet-18\citep{he2016resnet} truncated after its third residual stage (256 channels, stride 16) and projected to 384 channels by a frozen $1\times1$ convolution; inputs are mapped to $[0,1]$ and standardized with ImageNet statistics before the teacher, which outputs a $384\times4\times4$ map for a $64\times64$ input.

The student is a multi-scale patch description network (PDN) operating on $[-1,1]$ inputs, with three parallel branches (128 channels each) using kernel sizes $k \in \{4,6,8\}$, $2\times2$ average pooling with stride 2, and LeakyReLU (slope 0.1). The branches are cropped to a common size, concatenated, and fused by a $1\times1$ convolution to a $384\times25\times25$ output. A convolutional encoder-decoder with skip connections (three $3\times3$ stages with 64, 128, and 256 channels) is trained jointly to regress the teacher features but does not contribute to the anomaly score. Discrepancies are computed on the overlapping top-left $4\times4$ window of the teacher and student grids. With cropped features $T(x)$ and $S(x)$ ($C=384$ channels),
\begin{equation}
d(x)_{ij} = \frac{1}{C}\sum_{c=1}^{C}\bigl(T(x)_{cij} - S(x)_{cij}\bigr)^2 .
\end{equation}
Training uses a hard-feature loss: only the $k = \max(1, \lfloor 0.2\,hw \rfloor) = 3$ positions with the largest discrepancy contribute to the student loss $\mathcal{L}_{\mathrm{st}}$. The autoencoder loss $\mathcal{L}_{\mathrm{ae}}$ is the mean squared error between the teacher features and the autoencoder's bilinearly resized output. An out-of-distribution term regularizes the student on channel-permuted inputs (an independent permutation $\pi$ per sample),
\begin{equation}
\mathcal{L}_{\mathrm{ood}} = \Bigl(\overline{S(\pi(x))^{2}} + 10^{-8}\Bigr)^{-1} ,
\end{equation}
which pushes the student's feature energy on permuted inputs upward. The total objective is
\begin{equation}
\mathcal{L}_{\mathrm{EAD}} = \mathcal{L}_{\mathrm{st}} + \mathcal{L}_{\mathrm{ae}} + 0.1\,\mathcal{L}_{\mathrm{ood}} .
\end{equation}

\subsection{Generative Module: Diffusion-Based Reconstruction Detection}
\label{sec:met_ddpm}

The generative branch trains a DDPM on normal wafers only and scores anomalies by reconstruction error under partial diffusion.\citep{ho2020,wyatt2022} Forward diffusion uses a cosine schedule over $T=1000$ steps (offset $s=0.008$, $\beta_t$ clipped to $[10^{-4}, 0.9999]$)\citep{nichol2021}:
\begin{equation}
x_t = \sqrt{\bar{\alpha}_t}\,x_0 + \sqrt{1-\bar{\alpha}_t}\,\epsilon, \qquad \epsilon \sim \mathcal{N}(0, I), \qquad \bar{\alpha}_t = \prod_{s=1}^{t}(1-\beta_s) .
\end{equation}

The reverse process is a U-Net $\epsilon_\theta(x_t,t)$ with a 128-dimensional sinusoidal timestep embedding projected through an MLP to 320 dimensions. Residual blocks (group normalization, SiLU, $3\times3$ convolutions) operate at three resolutions ($64\times64$, $32\times32$, $16\times16$) with channel widths $[48,96,192]$ and two blocks per level; self-attention\citep{vaswani2017attention} with four heads is applied at $16\times16$. The bottleneck at $16\times16$ with 192 channels uses three residual blocks, rather than the two used elsewhere, to add capacity without changing the receptive field. The decoder mirrors the encoder, using skip-connection concatenation and transposed convolution upsampling. 

Training minimizes the simplified noise-prediction objective\citep{ho2020}
\begin{equation}
\mathcal{L}_{\mathrm{DDPM}} = \mathbb{E}_{x_0,\,t,\,\epsilon}\bigl\| \epsilon - \epsilon_\theta(x_t, t) \bigr\|_2^2 ,
\end{equation}
with $t$ drawn uniformly on $\{0,\ldots,T-1\}$, AdamW (learning rate $2\times10^{-4}$, weight decay $10^{-5}$), cosine annealing over 30 epochs, gradient clipping at 1.0, batch size 32 (630 updates total), and an exponential moving average of the weights (decay 0.9999) used for all subsequent sampling and scoring. Sampling and scoring both use deterministic DDIM updates ($\eta=0$).\citep{song2021} At each update, $\hat{x}_0 = (x_t - \sqrt{1-\bar{\alpha}_t}\,\epsilon_\theta(x_t,t))/\sqrt{\bar{\alpha}_t}$ is clipped to $[-1,1]$, and $x_{t'} = \sqrt{\bar{\alpha}_{t'}}\,\hat{x}_0 + \sqrt{1-\bar{\alpha}_{t'}}\,\epsilon_\theta(x_t,t)$. Unconditional generation uses 50 steps from $t=999$ to $t=0$. For scoring, a test image is forward-diffused once to $t^*=500$ (where $\bar{\alpha}_{500}=0.4921$) and denoised back with 50 DDIM steps to give the reconstruction $\hat{x}$. The anomaly map is
\begin{equation}
A_{\mathrm{DDPM}}(x)_{ij} = \frac{1}{3}\sum_{c=1}^{3}\bigl(x_{0,cij} - \hat{x}_{cij}\bigr)^2 ,
\end{equation}
computed over all three channels and all $64\times64$ positions, with the raw score $r_{\mathrm{DDPM}}(x)$ its spatial mean. Partial diffusion to $t^*=500$ preserves enough input structure that a normal wafer should reconstruct with low error while a distributional violation should not, even though both share low-level statistics such as die density.

\subsection{Score Calibration and Fusion}
\label{sec:met_fusion}

Raw scores are calibrated independently per module because they have unrelated units and dynamic ranges. The lower anchor $q_{\mathrm{lo}}$ is the 1st percentile of validation normal scores; the upper anchor $q_{\mathrm{hi}}$ is the 99th percentile of validation defective scores (falling back to the pooled distribution if too few defective validation samples are available). The calibrated score is
\begin{equation}
s(x) = \mathrm{clip}\!\left(\frac{r(x) - q_{\mathrm{lo}}}{q_{\mathrm{hi}} - q_{\mathrm{lo}} + 10^{-8}},\; 0,\; 1\right).
\end{equation}
This anchors 0 to a typical normal wafer and 1 to a clearly defective one for each module separately; calibrated scores are monotone transforms of the raw scores and are not probabilities. Per-module thresholds $\tau$ were set on the validation set by maximizing Youden's $J$ statistic ($\mathrm{TPR}-\mathrm{FPR}$); a wafer is labeled defective when $s(x) \geq \tau$.

The fused score is
\begin{equation}
s_{\mathrm{fused}}(x) = \alpha \, s_{\mathrm{EAD}}(x) + (1-\alpha)\, s_{\mathrm{DDPM}}(x),
\end{equation}
with $\alpha=0.5$ throughout this study. The fused threshold was recalibrated on the fused validation scores by the same criterion. Spatial maps are fused analogously after per-sample min-max normalization, as $\alpha\,\tilde{A}_{\mathrm{EAD}}(x) + (1-\alpha)\,\tilde{A}_{\mathrm{DDPM}}(x)$.

\subsection{Evaluation Protocol}
\label{sec:met_eval}

Threshold-free metrics (AUROC, average precision) were computed from calibrated scores, while threshold-dependent metrics (F1, Matthews correlation coefficient, balanced accuracy) were derived from predictions at the validation-selected threshold. AUROC and AP are reported together due to their divergence under class imbalance, with precision-recall often being more informative on skewed data \citep{saito2015pr}. MCC is noted for its lower susceptibility to imbalance-driven inflation compared to F1 or accuracy \citep{chicco2020mcc}. Uncertainty was quantified using a nonparametric bootstrap (1,000 resamples with replacement, with degenerate single-class resamples discarded, and 95\% confidence intervals from the 2.5th and 97.5th percentiles), keeping thresholds fixed \citep{efron1994}. Pairwise comparisons included a paired bootstrap on the AUROC difference (shared resample indices across detectors, with a two-sided empirical $p$-value computed as twice the smaller tail fraction, capped at 1), the DeLong test on paired test-set scores \citep{delong1988}, and the continuity-corrected McNemar test on paired thresholded predictions \citep{mcnemar1947}, where $\chi^2 = (|b-c|-1)^2/(b+c)$ with $b$ and $c$ as counts correct under only one detector. Fusion-weight sensitivity was assessed by sweeping $\alpha$ from 0 to 1 in 0.1 increments, re-estimating the fused threshold on the validation set at each step, with $\alpha=0.5$ fixed beforehand for this analysis. Finally, 500 test wafers were randomly drawn, their student-network feature maps were global-average-pooled to 384 dimensions, and the resulting vectors were projected to two dimensions with t-SNE (perplexity 40, 1,000 iterations, automatic learning rate) \citep{vandermaaten2008}.

\section{Results and Discussion}
\label{sec:results}

\subsection{Dataset Characteristics}
\label{sec:res_data}

 The WM-38K dataset contains 1 normal pattern, 8 single-defect patterns, and 29 mixed-defect patterns. The images are sized at 52×52 pixels \citep{9184890}. Thirty-five classes have exactly 1,000 wafers, with exceptions for Near-Full (C07, about 149), Random (C09, about 866), and Center+Edge-Loc+Scratch (C24, 2,000). Two and three-primitive mixtures account for 13,000 wafers, making up 81.0\% of defective wafers; wafers with at least two defect primitives are included. The pixel-wise defect probability map shows three spatial structures: a central hotspot near 0.40 from the Center and Donut patterns, a thin peripheral ring above 0.50 from the Edge-Ring and Edge-Loc patterns, and a faint arc in the lower half from the Scratch. The remaining die field sits near 0.25, reflecting random background failures across all classes. The normal class (C01) is not defect-free, containing sparse, random broken dies. The one-class boundary needs to separate structured patterns from a nonzero random background rather than just detect failing dies.

\subsection{Model Configuration and Training Dynamics}
\label{sec:res_training}

\begin{figure}[t]
  \centering
  \begin{subfigure}[b]{0.245\textwidth}
    \includegraphics[width=\textwidth]{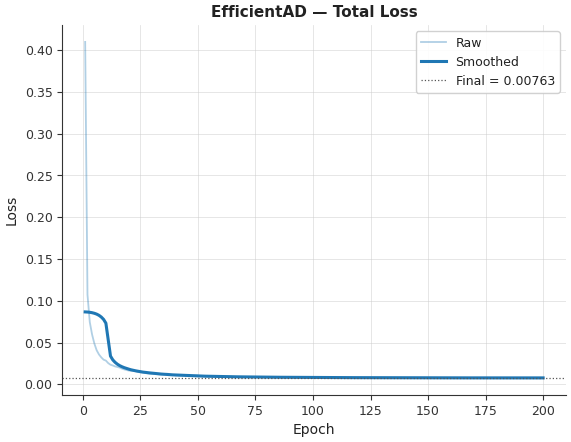}
    \caption{EfficientAD total}
  \end{subfigure}\hfill
  \begin{subfigure}[b]{0.245\textwidth}
    \includegraphics[width=\textwidth]{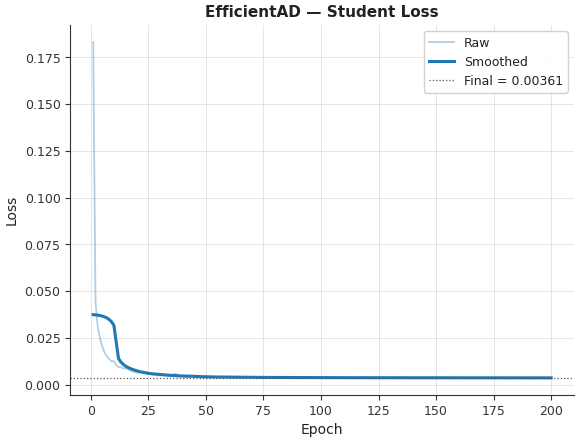}
    \caption{EfficientAD student}
  \end{subfigure}\hfill
  \begin{subfigure}[b]{0.245\textwidth}
    \includegraphics[width=\textwidth]{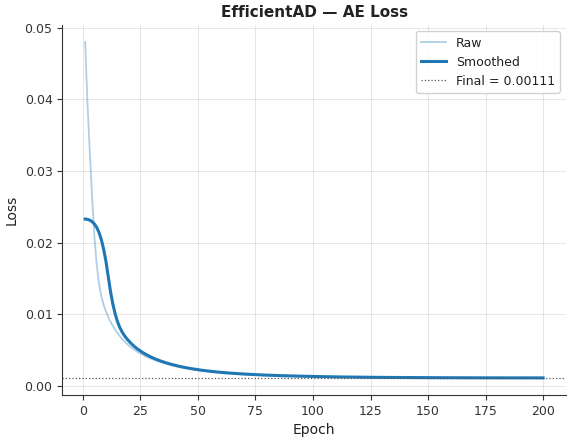}
    \caption{EfficientAD autoencoder}
  \end{subfigure}\hfill
  \begin{subfigure}[b]{0.245\textwidth}
    \includegraphics[width=\textwidth]{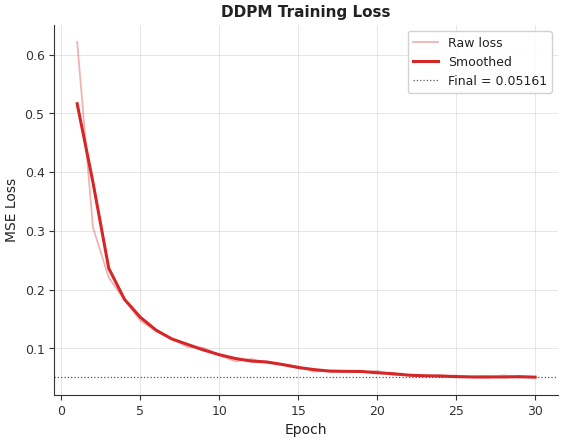}
    \caption{DDPM noise prediction}
  \end{subfigure}
  \caption{Training loss curves (raw and smoothed). Dotted lines mark final values.}
  \label{fig:training}
\end{figure}

Each epoch contains 21 minibatches of 32 normal wafers. EfficientAD receives 4,200 parameter updates over 200 epochs, while DDPM receives 630 updates over 30 epochs. EfficientAD converges early (Figure \ref{fig:training}), with its total loss dropping from 0.40984 to 0.01025 by epoch 50 (a 97.5\% reduction) and reaching 0.00763 at epoch 200. The final 150 epochs contribute only 25.6\% of the additional reduction. The student and autoencoder terms decrease by 98.0\% (0.18333 to 0.00361) and 97.7\% (0.04799 to 0.00111), respectively. The objective sums the student loss, the autoencoder loss, and the out-of-distribution (OOD) penalty, weighted at 0.1, with the residual 0.00291 assigned to the weighted OOD term, accounting for 38.1\% of the final objective. This indicates that the regularizer suppresses the student's off-manifold response throughout training. The DDPM loss drops from 0.62178 to 0.05161 (91.7\%) but continues to decline at epoch 30, with an additional 4.5\% decrease over the last five epochs. It receives 6.7 times fewer updates than EfficientAD.

\subsection{Detection Performance}
\label{sec:res_performance}

Table~\ref{tab:main} presents test-set performance, with Hybrid fusion leading in all metrics. Its AUROC of 0.9985 surpasses DDPM (0.9931) and EfficientAD (0.9799), with non-overlapping bootstrap intervals for AUROC, F1, and balanced accuracy between fusion and EfficientAD. Notably, the generative module outperforms the structural module across metrics, contrary to expectations for patch-level detectors on geometric defects. Average precision is saturated at 0.9998 for all detectors, and with a defective prevalence of 99.2\%, even random ranking yields an AP close to 0.992, limiting real differences and discriminative information \citep{saito2015pr}. Additionally, MCC remains low (0.278 to 0.432) despite F1 exceeding 0.96 for all models, as MCC penalizes errors on the minority normal class, which F1 does not address \citep{chicco2020mcc}. 

\begin{table}[t]
  \centering
  \caption{Test-set performance with 95\% bootstrap confidence intervals. Best values are in bold. Params count trainable weights only.}
  \label{tab:main}
  \small
  \setlength{\tabcolsep}{4pt}
  \begin{tabular}{lcccccc}
    \toprule
    Model & AUROC & AP & F1 & MCC & Bal.\ Acc. & Params (M) \\
    \midrule
    EfficientAD & 0.9799 & 0.9998 & 0.9604 & 0.2781 & 0.9288 & 5.21 \\
     & {\scriptsize[0.9728, 0.9866]} & {\scriptsize[0.9998, 0.9999]} & {\scriptsize[0.9582, 0.9624]} & {\scriptsize[0.2541, 0.3033]} & {\scriptsize[0.9072, 0.9473]} & \\
    DDPM & 0.9931 & 0.9999 & 0.9764 & 0.3620 & 0.9538 & 7.83 \\
     & {\scriptsize[0.9907, 0.9952]} & {\scriptsize[0.9999, 1.0000]} & {\scriptsize[0.9747, 0.9779]} & {\scriptsize[0.3310, 0.3920]} & {\scriptsize[0.9353, 0.9692]} & \\
    Fusion ($\alpha=0.5$) & \textbf{0.9985} & \textbf{1.0000} & \textbf{0.9830} & \textbf{0.4321} & \textbf{0.9800} & 13.04 \\
     & {\scriptsize[0.9978, 0.9991]} & {\scriptsize[1.0000, 1.0000]} & {\scriptsize[0.9817, 0.9843]} & {\scriptsize[0.3996, 0.4645]} & {\scriptsize[0.9721, 0.9843]} & \\
    \bottomrule
  \end{tabular}
\end{table}

\begin{figure}[t]
  \centering
  \begin{subfigure}[b]{0.50\textwidth}
    \includegraphics[width=\textwidth]{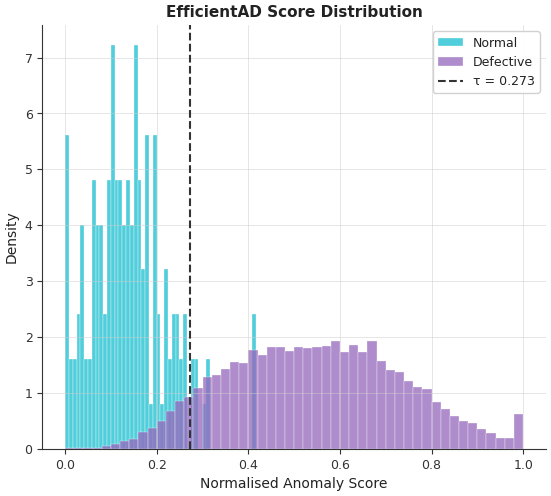}
    \caption{EfficientAD ($\tau = 0.273$)}
  \end{subfigure}\hfill
  \begin{subfigure}[b]{0.50\textwidth}
    \includegraphics[width=\textwidth]{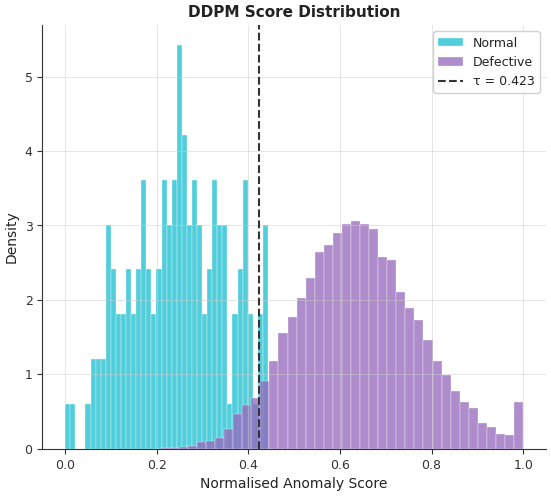}
    \caption{DDPM ($\tau = 0.423$)}
  \end{subfigure}
  \caption{Normalized test-set anomaly score distributions. Dashed lines mark validation-selected thresholds.}
  \label{fig:scores}
\end{figure}

Figure~\ref{fig:scores} explains the ranking. For EfficientAD, normal scores concentrate between 0 and 0.30, but the defective distribution has a long left tail that reaches about 0.10 and overlaps the normal mode at the threshold $\tau = 0.273$. The DDPM places its defective mode near 0.62 and, more importantly, its defective density vanishes below about 0.30 ($\tau = 0.423$). The DDPM's advantage therefore comes from compressing the lower tail of the defective distribution, not from pulling normal scores further down.

\begin{figure}[t]
  \centering
  \begin{subfigure}[b]{0.325\textwidth}
    \includegraphics[width=\textwidth]{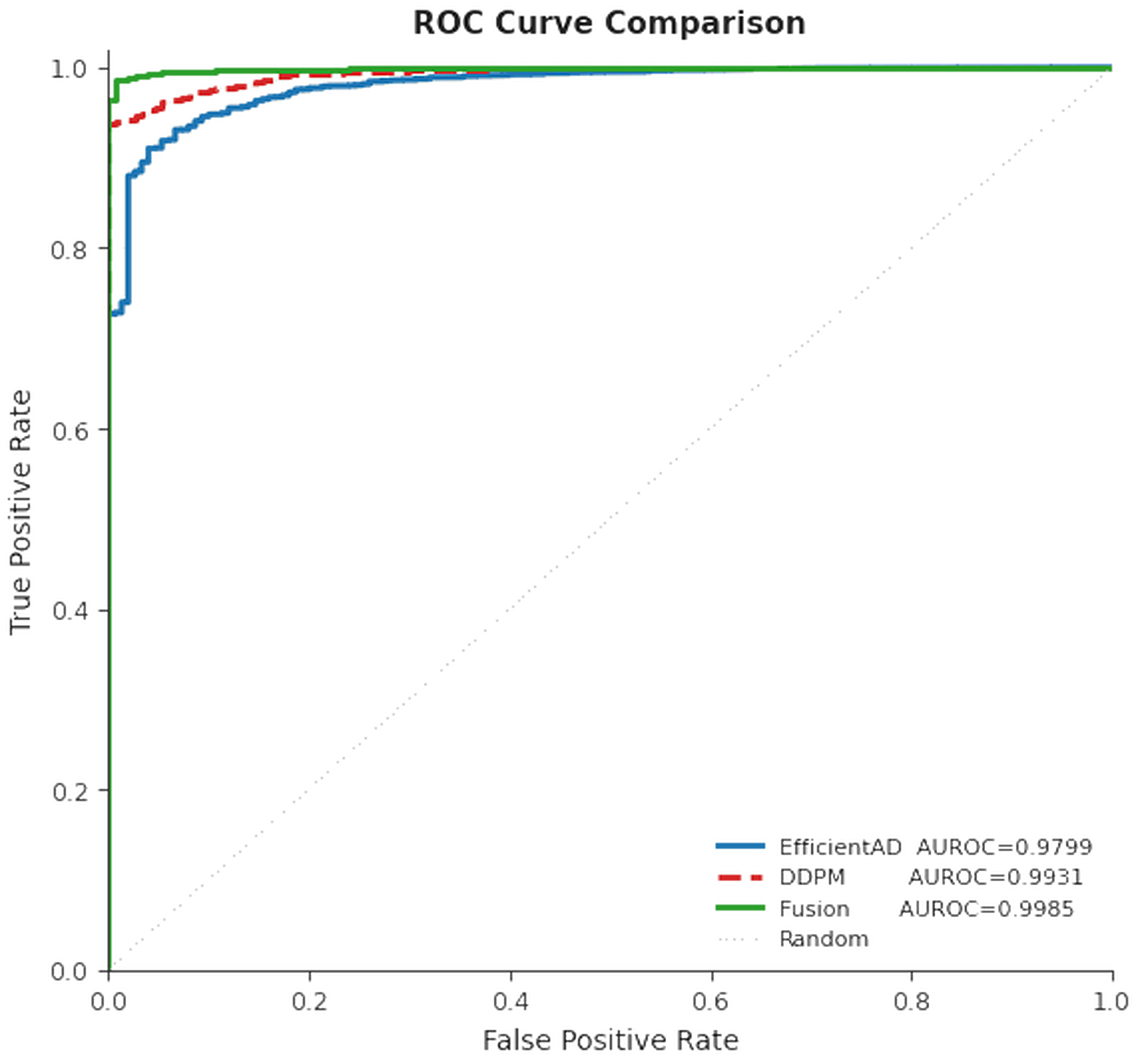}
    \caption{ROC}
  \end{subfigure}\hfill
  \begin{subfigure}[b]{0.325\textwidth}
    \includegraphics[width=\textwidth]{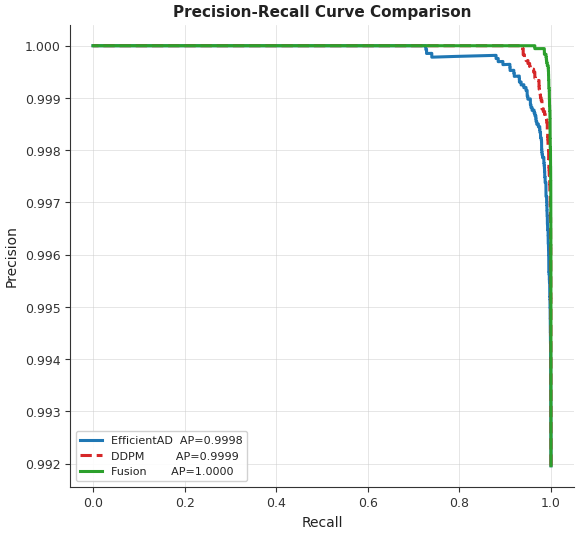}
    \caption{Precision-recall}
  \end{subfigure}\hfill
  \begin{subfigure}[b]{0.325\textwidth}
    \includegraphics[width=\textwidth]{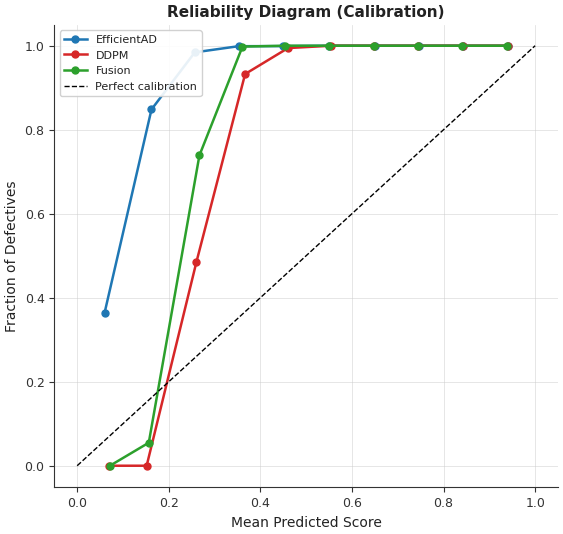}
    \caption{Reliability diagram}
  \end{subfigure}
  \caption{Three-way comparison of EfficientAD, DDPM, and hybrid fusion on the test set.}
  \label{fig:curves}
\end{figure}

Figure~\ref{fig:curves} compares the three detectors. The ROC curves separate most clearly in the high-specificity regime. At a false-positive rate near zero, fusion detects about 97\% of defective wafers, while DDPM detects about 94\% and EfficientAD about 73\%. This is crucial in production as every false alarm prompts manual review. The precision-recall curves reflect a similar trend. EfficientAD's precision drops near a recall of 0.73, DDPM's near 0.93, while fusion maintains a precision of 1.0 until a recall of about 0.97. In the reliability diagram, all curves lie above the diagonal, with scores being percentile-anchored ranks rather than probabilities, thus underestimating defect frequency. Post-hoc recalibration, such as isotonic regression, would be necessary for risk estimates. The lowest EfficientAD bin (mean score around 0.06) includes 36\% defective wafers, indicating left-tail contamination. In contrast, the DDPM and fusion bins with scores below 0.16 contain no defective wafers.

\subsection{Operating Point, Error Complementarity, and Significance}
\label{sec:res_errors}

Table~\ref{tab:confusion} presents the confusion counts at the selected validation thresholds. These counts correspond to the reported F1, MCC, and balanced accuracy under fixed class totals, and they reproduce all McNemar discordance counts in Table~\ref{tab:significance}. Fusion reduces the total error count from 1,412 (EfficientAD) and 852 (DDPM) to 618, and false alarms from 10 and 7 to one wafer. The table also clarifies the low MCC; of the 766 wafers labeled normal by fusion, only 149 are genuinely normal, resulting in a negative predictive value (NPV) of 19.5\%. This reflects the test composition rather than a model issue. Even at a 96.7\% true-positive rate, the 617 missed defects outnumber the 150 normal wafers by a factor of four. Thus, a "defective" verdict is nearly certain (PPV of 0.9999), while a "normal" verdict is unreliable given this prevalence.

\begin{table}[t]
  \centering
  \caption{Confusion counts and derived rates at the validation-selected thresholds (test set; 18,508 defective, 150 normal).}
  \label{tab:confusion}
  \small
  \setlength{\tabcolsep}{3.5pt}
  \begin{tabular}{lccccccccc}
    \toprule
    Model & TP & FN & TN & FP & Errors & TPR & TNR & PPV & NPV \\
    \midrule
    EfficientAD & 17,106 & 1,402 & 140 & 10 & 1,412 & 0.924 & 0.933 & 0.9994 & 0.091 \\
    DDPM        & 17,663 & 845   & 143 & 7  & 852   & 0.954 & 0.953 & 0.9996 & 0.145 \\
    Fusion      & \textbf{17,891} & \textbf{617} & \textbf{149} & \textbf{1} & \textbf{618} & \textbf{0.967} & \textbf{0.993} & \textbf{0.9999} & \textbf{0.195} \\
    \bottomrule
  \end{tabular}
\end{table}

Table~\ref{tab:significance} shows significant pairwise differences at $p<0.001$ across three tests: a paired bootstrap on the AUROC difference, the DeLong test for correlated ROC curves \citep{delong1988}, and the continuity-corrected McNemar test on thresholded predictions \citep{mcnemar1947}. The McNemar counts reveal why fusion works. EfficientAD and the DDPM disagree on 1,720 wafers but fail together on only 272. The DDPM correctly classifies 1,140 of the 1,412 EfficientAD errors (80.7\%), while EfficientAD correctly classifies 580 of the 852 DDPM errors (68.1\%). This low joint-error rate indicates diversity, under which outlier ensembles improve on their best member \citep{aggarwal2015}. Fusion repairs 895 EfficientAD errors with only 101 new ones and repairs 482 DDPM errors with 248 new ones. An oracle would make only 272 errors. Fixed-weight fusion (618 errors) captures 69.6\% of available headroom relative to EfficientAD and 40.3\% relative to the DDPM. The remaining 346 recoverable errors define the target for a sample-adaptive gating function.

\begin{table}[t]
  \centering
  \caption{Pairwise significance tests on the test set. For McNemar, ``A-only'' and ``B-only'' count wafers misclassified by one model alone; ``Both'' counts shared errors. EAD: EfficientAD. *** denotes $p<0.001$.}
  \label{tab:significance}
  \footnotesize
  \setlength{\tabcolsep}{3pt}
  \begin{tabular}{lcccccccc}
    \toprule
    & \multicolumn{2}{c}{Paired bootstrap} & \multicolumn{2}{c}{DeLong} & \multicolumn{4}{c}{McNemar} \\
    \cmidrule(lr){2-3}\cmidrule(lr){4-5}\cmidrule(lr){6-9}
    Comparison (A vs.\ B) & $\Delta$AUROC & 95\% CI & $z$ & $p$ & A-only & B-only & Both & $\chi^2$ \\
    \midrule
    Fusion vs.\ EAD & $+0.0184$ & [$+0.0123$, $+0.0256$] & $+5.335$ & $9.6\times10^{-8}$ & 101 & 895 & 517 & 631.37*** \\
    Fusion vs.\ DDPM        & $+0.0054$ & [$+0.0034$, $+0.0078$] & $+5.124$ & $3.0\times10^{-7}$ & 248 & 482 & 370 & 74.37*** \\
    EAD vs.\ DDPM   & $-0.0130$ & [$-0.0209$, $-0.0056$] & $-3.423$ & $6.2\times10^{-4}$ & 1,140 & 580 & 272 & 181.68*** \\
    \bottomrule
  \end{tabular}
\end{table}

\subsection{Sensitivity to the Fusion Weight}
\label{sec:res_alpha}

\begin{figure}[t]
  \centering
  \includegraphics[width=0.55\textwidth]{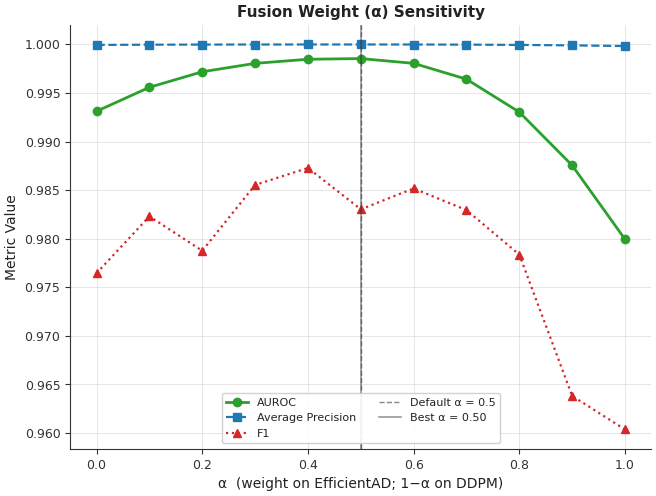}
  \caption{Test-set AUROC, AP, and F1 as a function of the fusion weight $\alpha$ ($\alpha = 0$: DDPM only; $\alpha = 1$: EfficientAD only).}
  \label{fig:alpha}
\end{figure}

Figure~\ref{fig:alpha} shows the fusion weight $\alpha$ for EfficientAD from 0 to 1. AUROC is concave in $\alpha$, peaking at $\alpha = 0.5$ (0.99854), with $\alpha = 0.4$ close behind (0.99847). Weights from 0.1 to 0.7 outperform both single modules. The curve is asymmetric: moving from $\alpha = 0.5$ to pure EfficientAD reduces AUROC by 0.0186, whereas transitioning to pure DDPM reduces it by 0.0054, reflecting the stronger performance of DDPM. Even a 10\% EfficientAD weight increases AUROC from 0.9931 to 0.9956. F1 fluctuates between 0.9788 and 0.9873 for $\alpha$ from 0.1 to 0.6, peaking at $\alpha = 0.4$, with fluctuations driven by threshold noise due to limited validation normals. AP remains at or above 0.99983 for all $\alpha$. The default $\alpha = 0.5$ was established before this sweep, making the results in Table~\ref{tab:main} not tuned on test data; the sweep itself uses the test set for sensitivity analysis.

\subsection{Spatial Anomaly Maps and Generative Fidelity}
\label{sec:res_spatial}

\begin{figure}[t]
  \centering
  \begin{subfigure}[b]{0.8\textwidth}
    \includegraphics[width=\textwidth]{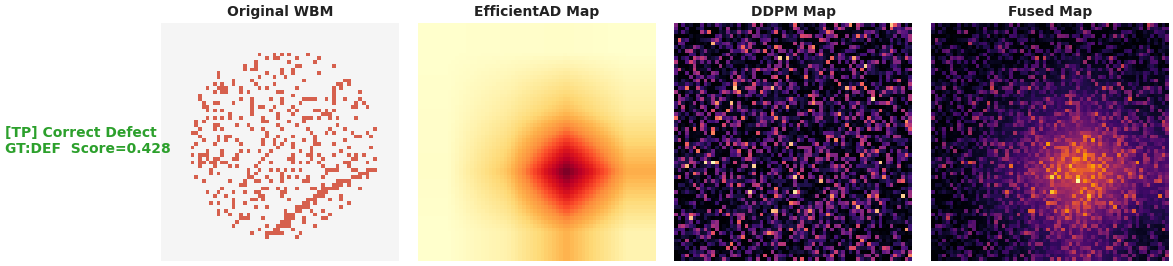}
  \end{subfigure}\\[-2pt]
  \begin{subfigure}[b]{0.8\textwidth}
    \includegraphics[width=\textwidth]{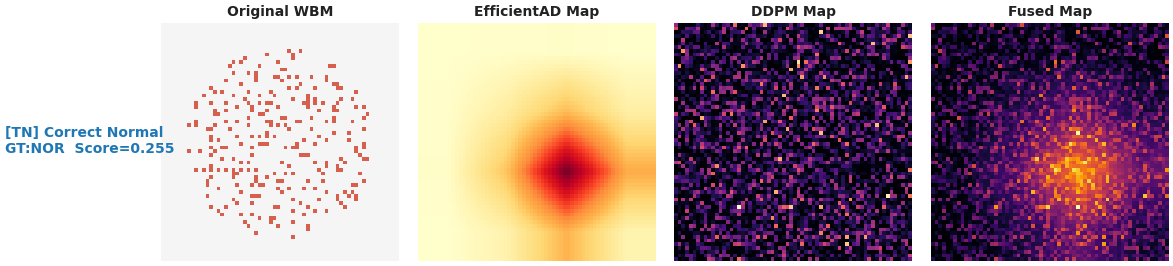}
  \end{subfigure}\\[-2pt]
  \begin{subfigure}[b]{0.8\textwidth}
    \includegraphics[width=\textwidth]{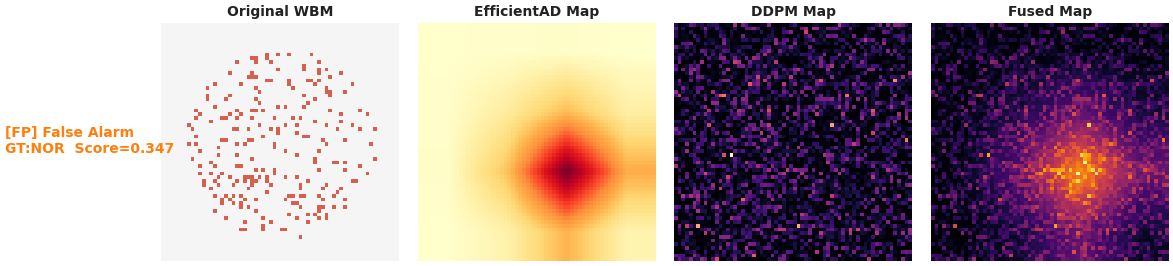}
  \end{subfigure}\\[-2pt]
  \begin{subfigure}[b]{0.8\textwidth}
    \includegraphics[width=\textwidth]{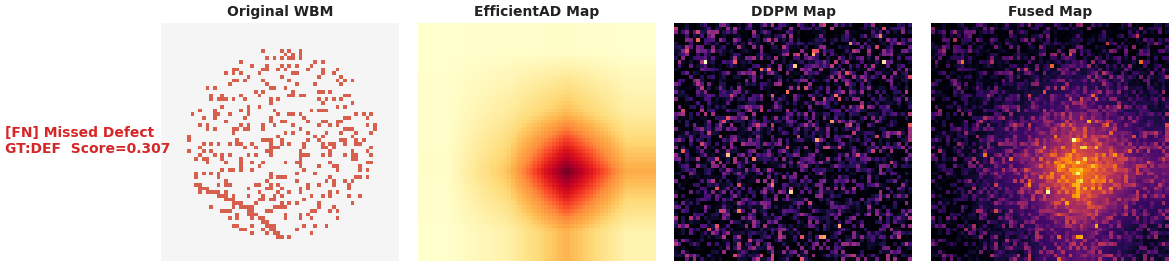}
  \end{subfigure}
  \caption{Spatial anomaly map decomposition for one true-positive, true-negative, false-positive, and false-negative test wafer (top to bottom). Columns: original wafer map, EfficientAD map, DDPM reconstruction-error map, and fused map. Scores are fused image-level scores.}
  \label{fig:spatial}
\end{figure}

Figure~\ref{fig:spatial} decomposes one test wafer from each outcome category. The EfficientAD map shows a diamond-shaped hotspot slightly below and to the right of the wafer center, which is an architectural artifact. With a $64\times64$ input, the truncated ResNet-18 teacher generates a $4\times4$ feature grid (stride 16), while the student produces a $25\times25$ grid. The discrepancy is calculated over the overlapping $4\times4$ window, which does not align with the teacher cells. This $4\times4$ map is bilinearly upsampled by a factor of 16. It cannot resolve 52-die defect geometry but separates the classes based on the global mass of broken dies, lacking pixel-level localization. The DDPM map is unstructured and spreads across the blank padding region due to a single random noise draw during the forward diffusion to $t^* = 500$, followed by averaging over three channels. This results in a stable image-level score, but the per-pixel map lacks localization. The fused map combines the EfficientAD hotspot with the DDPM texture, improving image-level decisions without enhancing localization. The error cases highlight the hardest defect type. Both the true positive (score 0.428) and false negative (score 0.307) are Scratch wafers, with the missed one being a thin arc along the lower-left edge. The false positive is a normal wafer with a score of 0.347. Thus, the fused threshold lies between 0.307 and 0.347. Thin, low-density scratches contribute few broken dies compared to random background, causing detectors that score global defect mass to rate them similarly to normal wafers.

\begin{figure}[t]
  \centering
  \includegraphics[width=0.72\textwidth]{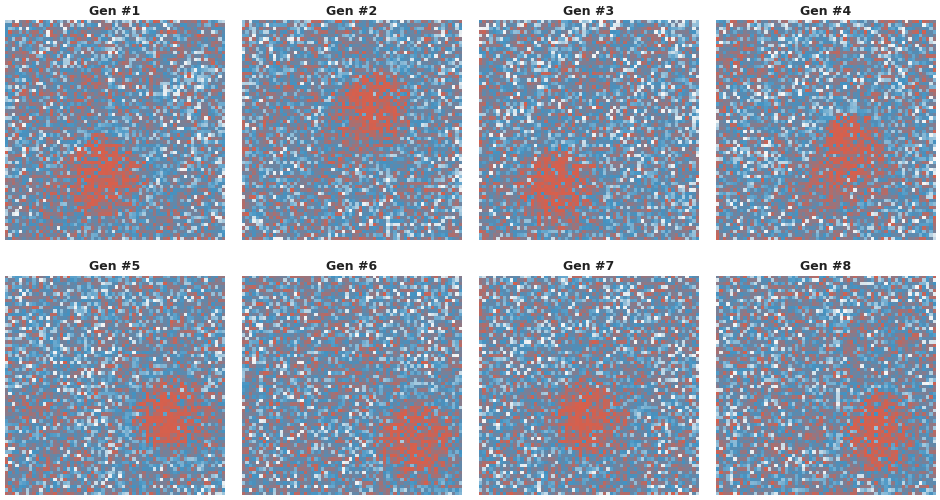}
  \caption{Unconditional DDIM samples (50 steps) from the DDPM trained on 700 normal wafers (broken-die channel shown).}
  \label{fig:generated}
\end{figure}

Figure~\ref{fig:generated} presents unconditional samples from a trained DDPM using 50-step deterministic DDIM sampling, achieving 11.14 iterations per second on the T4 (approximately 4.5 s per 50-step pass). The samples fill the $64\times64$ canvas with dense die texture but lack a circular wafer boundary, each displaying a single blob-like cluster of broken dies at random positions. The model captures local die statistics and cluster priors but fails on global wafer geometry, given only 630 updates across 700 training images. Nevertheless, it achieves an AUROC of 0.9931 as a detector. This performance is attributed to the scoring timestep; at $t^* = 500$, the noisy input retains 70\% of its signal amplitude, making the reverse process rely on local denoising rather than global synthesis. Thus, the generative module functions as a learned local density prior. Its detection ability is separate from generative fidelity, and the global "logical anomaly" reasoning typically associated with diffusion models does not drive its performance on this benchmark.

\section{Conclusion}
\label{sec:conclusion}
 
This study developed a hybrid one-class framework for wafer bin map anomaly
detection that pairs a patch-based student--teacher detector (EfficientAD) with
a diffusion-based reconstruction detector (DDPM) and fuses their calibrated
scores through a fixed convex weight. The framework was trained exclusively on
700 normal wafers from WM-38K, evaluated on 18,658 held-out wafers, and examined
not only for their detection accuracy but also for the behavior of the fusion
mechanism itself. Fixed equal-weight fusion outperformed both constituent detectors
on every metric. It raised AUROC to 0.9985, compared with 0.9931 for the DDPM and
0.9799 for EfficientAD, reduced the total error count to 618, and limited false
alarms to a single normal wafer. All pairwise differences were significant at
$p < 0.001$ under the paired bootstrap, DeLong, and McNemar tests. Fusion did not
degrade performance for any interior weight $0.1 \le \alpha \le 0.7$; degradation
appeared only as the weight approached pure EfficientAD. The gain stems from
error complementarity, since the two modules disagreed on 1,720 wafers but failed
together on only 272. Fixed weighting, however, captured only 69.6\% of the
oracle headroom relative to EfficientAD and 40.3\% relative to the DDPM, leaving
346 recoverable errors unexploited. The two modules produce qualitatively different score geometries.
EfficientAD assigns defective wafers a long left tail that overlaps the normal
mode, whereas the DDPM compresses the lower tail of the defective distribution
so that its density vanishes below a calibrated score of about 0.30. The fused
score inherits this clean lower tail while adding structural evidence. The
relationship between the fusion weight and AUROC is not monotonic; it is concave
and asymmetric, peaking at $\alpha = 0.5$ and penalizing reliance on the weaker
module far more than reliance on the stronger one. All three calibrated scores
lie above the diagonal of the reliability diagram, confirming that they are
percentile-anchored ranks rather than probabilities. The inverted class balance of WM-38K saturates the average precision
($\geq 0.9998$) and F1 ($> 0.96$), whereas MCC (0.278--0.432) and the negative
predictive value of the fused detector (19.5\%) reveal that ``normal'' verdicts
remain unreliable. Both modules effectively score global broken-die mass: the
EfficientAD map is an architectural artifact of a $4 \times 4$ teacher grid, and
the DDPM map is spatially unstructured. The DDPM detects well despite generating
samples that lack wafer geometry, because partial diffusion to $t^{*} = 500$
conditions reconstruction strongly on the input and reduces it to local
denoising. Consequently, thin, low-density scratches are the dominant residual
failure mode. 

This study has several limitations. WM-38K inverts the industrial base rate: 97.4\% of all wafers and 99.2\% of the test partition are defective, leaving only 150 normal wafers in each validation and test partition. All experiments use a single dataset with a native resolution of $52 \times 52$ and lack cross-dataset or cross-fab validation. The comparison includes only the constituent modules and their fusion; established one-class detectors were not evaluated under the same protocol. The structural module is limited in its ability to adapt to small inputs. Future work could include sample-adaptive fusion. A gating function using signals such as inter-module score disagreement or local defect density could target the 346 errors that remain unrecovered. To remain consistent with the one-class setting, the gate should rely on unsupervised cues or a separate held-out partition. Evaluation under realistic prevalence is important. Benchmarks like WM-811K with larger normal pools would stabilize threshold selection and allow reporting of operationally relevant quantities, including the false-alarm rate at a fixed recall. Conformal calibration could provide guaranteed control over the false-alarm rate on normal wafers. 

\section*{Abbreviations}
\noindent
AD, anomaly detection;
AE, autoencoder;
AP, average precision;
AUROC, area under the receiver operating characteristic curve;
CI, confidence interval;
DDIM, denoising diffusion implicit model;
DDPM, denoising diffusion probabilistic model;
EMA, exponential moving average;
MCC, Matthews correlation coefficient;
OOD, out-of-distribution;
PDN, patch description network;
ROC, receiver operating characteristic;
t-SNE, t-distributed stochastic neighbor embedding;
WBM, wafer bin map.
 
\section*{Author Contributions}
Limon Bin Hossain: conceptualization, methodology, software, formal analysis, writing -- original draft.
Md. Sadib Rahman Ananta: writing -- review and editing, validation, visualization.

\section*{Funding Sources}
This research did not receive any specific grant from funding agencies in the public, commercial, or not-for-profit sectors.

\section*{Notes}
The authors declare no competing financial interest.
\section*{Declaration of generative AI and AI-assisted technologies in the writing process}
	During the preparation of this work, the author(s) used QuilBot Premium /Grammarly Premium/ChatGPT/Claude Premium to improve the quality of the writing and check for any grammatical errors. After using this tool/service, the author(s) reviewed and edited the content as needed and take(s) full responsibility for the content of the publication.
\section*{Acknowledgments}
The authors thank the maintainers of the WM-38K (MixedWM38) dataset for making it publicly available for anomaly detection research.

\bibliography{ref}

@article{nakazawa2018,
  author  = {Nakazawa, T. and Kulkarni, D. V.},
  title   = {Wafer Map Defect Pattern Classification and Image Retrieval Using Convolutional Neural Network},
  journal = {IEEE Trans. Semicond. Manuf.},
  year    = {2018},
  volume  = {31},
  number  = {2},
  pages   = {309--314},
  doi     = {10.1109/TSM.2018.2795466}
}

@article{wang2020,
  author  = {Wang, J. and Xu, C. and Yang, Z. and Zhang, J. and Li, X.},
  title   = {Deformable Convolutional Networks for Efficient Mixed-Type Wafer Defect Pattern Recognition},
  journal = {IEEE Trans. Semicond. Manuf.},
  year    = {2020},
  volume  = {33},
  number  = {4},
  pages   = {587--596},
  doi     = {10.1109/TSM.2020.3020985}
}

@inproceedings{bergmann2020,
  author    = {Bergmann, P. and Fauser, M. and Sattlegger, D. and Steger, C.},
  title     = {Uninformed Students: Student-Teacher Anomaly Detection with Discriminative Latent Embeddings},
  booktitle = {Proceedings of the IEEE/CVF Conference on Computer Vision and Pattern Recognition (CVPR)},
  year      = {2020},
  pages     = {4183--4192},
  url       = {https://arxiv.org/abs/1911.02357}
}

@inproceedings{defard2021,
  author    = {Defard, T. and Setkov, A. and Loesch, A. and Audigier, R.},
  title     = {{PaDiM}: A Patch Distribution Modeling Framework for Anomaly Detection and Localization},
  booktitle = {International Conference on Pattern Recognition Workshops (ICPR-W)},
  year      = {2021},
  pages     = {475--489},
  url       = {https://arxiv.org/abs/2011.08785}
}

@inproceedings{dengli2022,
  author    = {Deng, H. and Li, X.},
  title     = {Anomaly Detection via Reverse Distillation from One-Class Embedding},
  booktitle = {Proceedings of the IEEE/CVF Conference on Computer Vision and Pattern Recognition (CVPR)},
  year      = {2022},
  pages     = {9737--9746},
  url       = {https://arxiv.org/abs/2201.10703}
}

@inproceedings{batzner2024,
  author    = {Batzner, K. and Heckler, L. and K{\"o}nig, R.},
  title     = {{EfficientAD}: Accurate Visual Anomaly Detection at Millisecond-Level Latencies},
  booktitle = {Proceedings of the IEEE/CVF Winter Conference on Applications of Computer Vision (WACV)},
  year      = {2024},
  pages     = {128--138},
  url       = {https://arxiv.org/abs/2303.14535}
}

@inproceedings{ho2020,
  author    = {Ho, J. and Jain, A. and Abbeel, P.},
  title     = {Denoising Diffusion Probabilistic Models},
  booktitle = {Advances in Neural Information Processing Systems (NeurIPS)},
  year      = {2020},
  volume    = {33},
  pages     = {6840--6851},
  url       = {https://arxiv.org/abs/2006.11239}
}

@inproceedings{nichol2021,
  author    = {Nichol, A. Q. and Dhariwal, P.},
  title     = {Improved Denoising Diffusion Probabilistic Models},
  booktitle = {Proceedings of the 38th International Conference on Machine Learning (ICML)},
  year      = {2021},
  pages     = {8162--8171},
  url       = {https://arxiv.org/abs/2102.09672}
}

@inproceedings{song2021,
  author    = {Song, J. and Meng, C. and Ermon, S.},
  title     = {Denoising Diffusion Implicit Models},
  booktitle = {International Conference on Learning Representations (ICLR)},
  year      = {2021},
  url       = {https://arxiv.org/abs/2010.02502}
}

@inproceedings{wyatt2022,
  author    = {Wyatt, J. and Leach, A. and Schmon, S. M. and Willcocks, C. G.},
  title     = {{AnoDDPM}: Anomaly Detection with Denoising Diffusion Probabilistic Models Using Simplex Noise},
  booktitle = {IEEE/CVF Conference on Computer Vision and Pattern Recognition Workshops (CVPRW)},
  year      = {2022},
  pages     = {650--656},
  url       = {https://arxiv.org/abs/2201.08268}
}

@article{aggarwal2015,
  author  = {Aggarwal, C. C. and Sathe, S.},
  title   = {Theoretical Foundations and Algorithms for Outlier Ensembles},
  journal = {ACM SIGKDD Explor. Newsl.},
  year    = {2015},
  volume  = {17},
  number  = {1},
  pages   = {24--47},
  doi     = {10.1145/2830544.2830549}
}

@book{efron1994,
  author    = {Efron, B. and Tibshirani, R. J.},
  title     = {An Introduction to the Bootstrap},
  publisher = {Chapman \& Hall/CRC},
  address   = {New York},
  year      = {1994},
  doi       = {10.1201/9780429246593}
}

@article{mcnemar1947,
  author  = {McNemar, Q.},
  title   = {Note on the Sampling Error of the Difference between Correlated Proportions or Percentages},
  journal = {Psychometrika},
  year    = {1947},
  volume  = {12},
  number  = {2},
  pages   = {153--157},
  doi     = {10.1007/BF02295996}
}

@article{vandermaaten2008,
  author  = {van der Maaten, L. and Hinton, G.},
  title   = {Visualizing Data Using t-{SNE}},
  journal = {J. Mach. Learn. Res.},
  year    = {2008},
  volume  = {9},
  pages   = {2579--2605},
  url     = {https://www.jmlr.org/papers/v9/vandermaaten08a.html}
}

@article{chicco2020mcc,
  author  = {Chicco, D. and Jurman, G.},
  title   = {The Advantages of the Matthews Correlation Coefficient ({MCC}) over {F1} Score and Accuracy in Binary Classification Evaluation},
  journal = {BMC Genomics},
  year    = {2020},
  volume  = {21},
  pages   = {6},
  doi     = {10.1186/s12864-019-6413-7}
}

@article{saito2015pr,
  author  = {Saito, T. and Rehmsmeier, M.},
  title   = {The Precision-Recall Plot Is More Informative than the {ROC} Plot When Evaluating Binary Classifiers on Imbalanced Datasets},
  journal = {PLOS ONE},
  year    = {2015},
  volume  = {10},
  number  = {3},
  pages   = {e0118432},
  doi     = {10.1371/journal.pone.0118432}
}

@article{delong1988,
  author  = {DeLong, E. R. and DeLong, D. M. and Clarke-Pearson, D. L.},
  title   = {Comparing the Areas under Two or More Correlated Receiver Operating Characteristic Curves: A Nonparametric Approach},
  journal = {Biometrics},
  year    = {1988},
  volume  = {44},
  number  = {3},
  pages   = {837--845},
  doi     = {10.2307/2531595}
}

@inproceedings{he2016resnet,
  author    = {He, K. and Zhang, X. and Ren, S. and Sun, J.},
  title     = {Deep Residual Learning for Image Recognition},
  booktitle = {Proceedings of the IEEE Conference on Computer Vision and Pattern Recognition (CVPR)},
  year      = {2016},
  pages     = {770--778}
}

@inproceedings{vaswani2017attention,
  author    = {Vaswani, A. and Shazeer, N. and Parmar, N. and Uszkoreit, J. and Jones, L. and Gomez, A. N. and Kaiser, {\L}. and Polosukhin, I.},
  title     = {Attention Is All You Need},
  booktitle = {Advances in Neural Information Processing Systems (NeurIPS)},
  year      = {2017},
  volume    = {30},
  pages     = {5998--6008}
}

@inproceedings{batzner2024efficientad,
  title={Efficientad: Accurate visual anomaly detection at millisecond-level latencies},
  author={Batzner, Kilian and Heckler, Lars and K{\"o}nig, Rebecca},
  booktitle={2024 IEEE/CVF Winter Conference on Applications of Computer Vision (WACV)},
  pages={127--137},
  year={2024},
  organization={IEEE}
}

@ARTICLE{9184890,
  author={Wang, Junliang and Xu, Chuqiao and Yang, Zhengliang and Zhang, Jie and Li, Xiaoou},
  journal={IEEE Transactions on Semiconductor Manufacturing}, 
  title={Deformable Convolutional Networks for Efficient Mixed-Type Wafer Defect Pattern Recognition}, 
  year={2020},
  volume={33},
  number={4},
  pages={587-596},
  doi={10.1109/TSM.2020.3020985}}

\end{document}